\documentclass{article}
\usepackage[utf8]{inputenc}
\usepackage{authblk}
\usepackage{setspace}
\usepackage[margin=1.25in]{geometry}
\usepackage{graphicx}
\usepackage{gensymb}
\graphicspath{ {./figures/} }
\usepackage{subcaption}
\usepackage{amsmath}
\usepackage{lineno}
\usepackage{xcolor}
\usepackage{tikz}
\usepackage{multirow}
\usepackage{booktabs}
\usepackage{adjustbox}
\usepackage{subcaption}
\usepackage[colorlinks=true, linkcolor=green, citecolor=green, urlcolor=green]{hyperref}
\usepackage[
  style=nejm,
  citestyle=numeric-comp,
  sorting=none,
  backref=true
]{biblatex}
\title{Enabling Plant Phenotyping in Weedy Environments using Multi-Modal Imagery via Synthetic and Generated Training Data}

\author[1]{Earl Ranario}
\author[1]{Ismael Mayanja}
\author[1]{Heesup Yun}
\author[2]{Brian N. Bailey}
\author[1*]{J. Mason Earles}
\affil[1]{Biological Systems Engineering, UC Davis, Davis, USA.}
\affil[2]{Plant Sciences, UC Davis, Davis, USA.}
\affil[*]{Address correspondence to: jmearles@ucdavis.edu}

\date{}

\begin{document}

\maketitle

%%%%%% Abstract %%%%%%
Accurate plant segmentation in thermal imagery remains a significant challenge for high throughput field phenotyping, particularly in outdoor environments where low contrast between plants and weeds and frequent occlusions hinder performance. To address this, we present a framework that leverages synthetic RGB imagery, a limited set of real annotations, and GAN-based cross-modality alignment to enhance semantic segmentation in thermal images. We trained models on 1,128 synthetic images containing complex mixtures of crop and weed plants in order to generate image segmentation masks for crop and weed plants. We additionally evaluated the benefit of integrating as few as five real, manually segmented field images within the training process using various sampling strategies. When combining all the synthetic images with a few labeled real images, we observed a maximum relative improvement of 22\% for the weed class and 17\% for the plant class compared to the full real-data baseline. Cross-modal alignment was enabled by translating RGB to thermal using CycleGAN-turbo, allowing robust template matching without calibration. Results demonstrated that combining synthetic data with limited manual annotations and cross-domain translation via generative models can significantly boost segmentation performance in complex field environments for multi-model imagery.

%%%%%% Main Text %%%%%%

\section{Introduction}

\begin{figure}[h]
    \centering
    \includegraphics[width=1.0\linewidth]{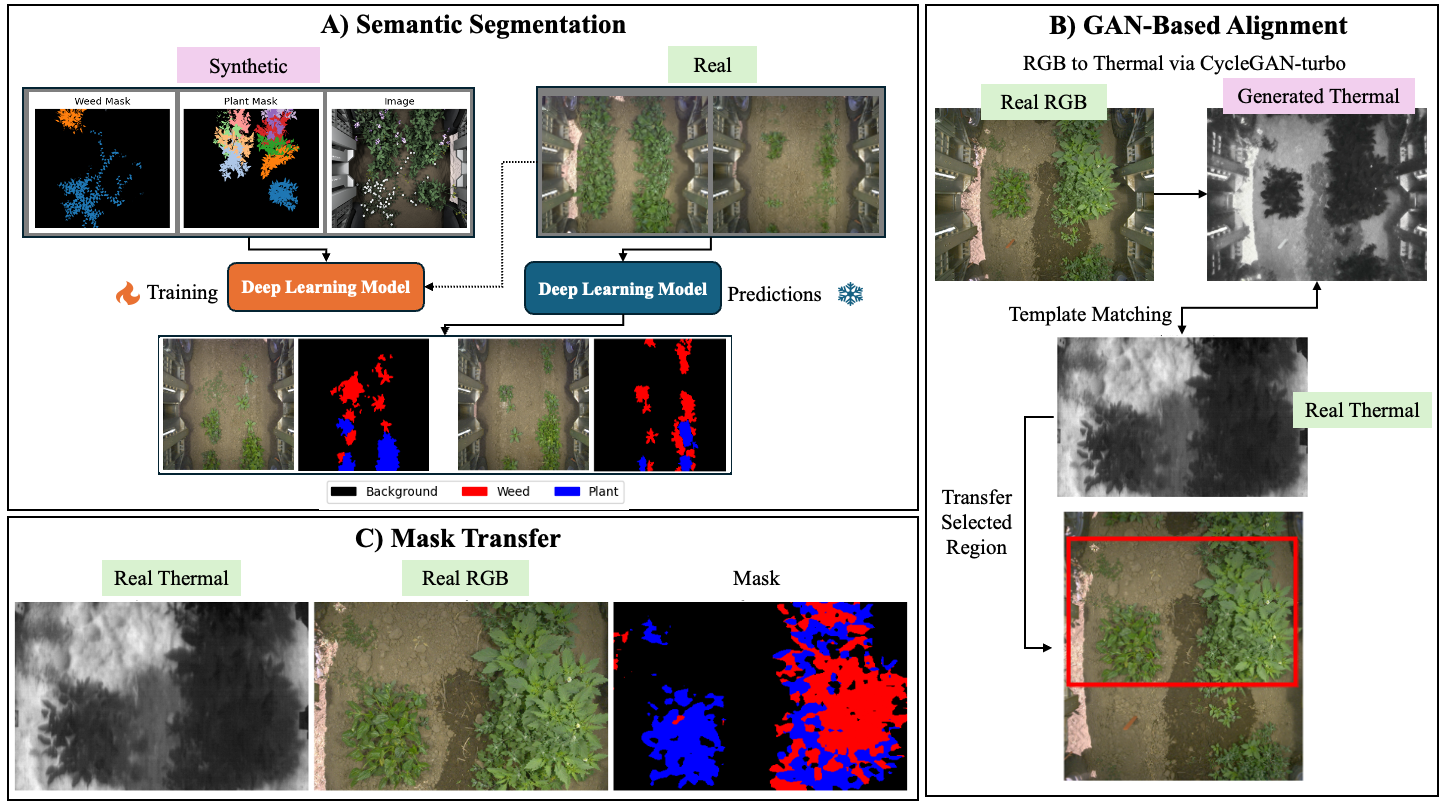}
    \caption{Overall framework to extract thermal values from the plant, excluding weeds and background. \textit{Part A)} The training of the semantic segmentation model using the synthetic images (including any real images if any). \textit{Part B)} The template matching of Real RGB and Real Thermal via CycleGAN-turbo. \textit{Part C)} Transfer prediction masks using selected region.}
    \label{fig:method}
\end{figure}

Open-field plant phenotyping faces significant challenges in separating target crops from complex backgrounds. In weedy field environments, plants often overlap with weeds and other clutter, causing occlusions and visual confusion in imagery \cite{cakic_evaluating_2025}. Crops and weeds frequently share similar colors, shapes, and textures, making it difficult for vision algorithms to distinguish between them \cite{ilyas_overcoming_2023, t_performance_2019, khan_ced-net_2020}. These issues lead to missed detections or false positives when using conventional segmentation approaches in the field. This difficulty is further apparent in non-RGB imaging modalities, such as thermal infrared. In thermal images, plant and soil temperatures can be similar under certain conditions, yielding low contrast between vegetation and background \cite{liu_recognition_2020}. Traditional methods for thermal segmentation often involve thresholding temperature histograms or watershed algorithms to isolate canopy pixels \cite{katz_how_2023}. In practice, researchers have found it advantageous to leverage co-registered RGB images to guide thermal image segmentation (e.g., masking the thermal image with an RGB-derived plant mask) \cite{rud_crop_2014, osroosh_economical_2018, zhou_ground-based_2022}. Overall, accurately segmenting individual plants in weedy, outdoor conditions, especially in modalities beyond visible light, remains an open problem.

Environmental factors further complicate the differentiation of weeds and crops of interest in thermal images. Sunlight exposure, wind, and moisture can alter plant temperatures in the field. A prior study observed that weeds just a few feet apart can register significantly different temperatures if one is sunlit and another is shaded \cite{eide_uav-assisted_2021}. This spatial temperature variability means the same weed species might appear hotter or cooler purely due to microclimate, not because they are biologically different. Another limitation is that different plant species under the same management conditions often have similar thermal signatures. For example, if a crop and a weed are both well-watered and not under stress, their canopy temperatures may be similar, making it difficult to distinguish crops from weeds \cite{shamshiri_sensing_2024}. In mixed vegetation, the overlapping thermal signals can confuse algorithms, leading to a higher rate of false detections. Zamani et al. \cite{zamani_earlylate_2023} created a dataset of paired visible and thermal UAV images in rice paddies to classify rice plants versus weeds \cite{zamani_earlylate_2023}. Their system achieved improved accuracy when combining both modalities in distinguishing weeds from crops by using a late-fusion neural network that combined features from both the visible and thermal images. 

Robust plant segmentation is a critical prerequisite for downstream phenotyping analyses. Masks of each plant allow computation of plant metrics such as canopy temperature, which in turn enable physiological traits to be inferred. For example, leaf and canopy temperatures are well-known proxies for transpiration and water status. Thermal imaging has been used to monitor drought response, as plant temperature correlates with transpiration rate and stomatal conductance \cite{mertens_monitoring_2023}. Additionally, thermal measurements are used to  monitor stomatal closure of grapevines \cite{jones_use_2002} and also wheat \cite{noauthor_physiological_nodate}. However, any background pixels mistakenly included in a plant's mask can skew these measurements. Even a small fraction of soil or residue mixed into the canopy temperature calculation can drastically alter indices like the crop water stress index (CWSI), sometimes producing non-physical values \cite{katz_how_2023}. 

Although modern machine learning tools have allowed for high-throughput prediction of plant traits from imagery data, the performance of these tools is limited by the availability of labeled data for model training \cite{jiang2020}. There is an emphasis on expanding public image datasets for agricultural tasks, but this alone may not adequately address the complexities of specific scenarios \cite{lu2020}. For instance, models trained in one domain may not generalize well to another due to differences in lighting, camera angle, or plant species. The general approach is to label new data tailored to specific domains, but this process is costly and time-consuming \cite{li_label-efficient_2023}. Manual labeling of image segmentation masks can be exceptionally time-consuming, as it requires careful identification and annotation of regions of interest in the images. Furthermore, accurate manual segmentation can be subject to human error, as it can be difficult to visually discern between object classes in images. However, by leveraging underlying representational knowledge encoded in a model that is trained on many data points, we can reduce the volume of real data required while still achieving competitive performance.

Our study aims to improve the segmentation accuracy of thermal images for cowpea plants grown in weedy, open-field conditions with a mixture of synthetic and real RGB images, as seen in Figure \ref{fig:phenology_grid}. The proposed approach, summarized in Figure \ref{fig:method}, combines synthetic image generation, domain adaptation, and cross-modality alignment to reduce manual annotation burdens and improve segmentation quality in multi-modal datasets. By tackling segmentation in these challenging conditions, we seek to enable more reliable extraction of plant temperature, ultimately advancing high-throughput field phenotyping in real-world, weedy fields. Although this work focuses on plant and weed segmentation in thermal imagery, the approach is generalizable to other imaging modalities with low object class contrast. Our contributions include:

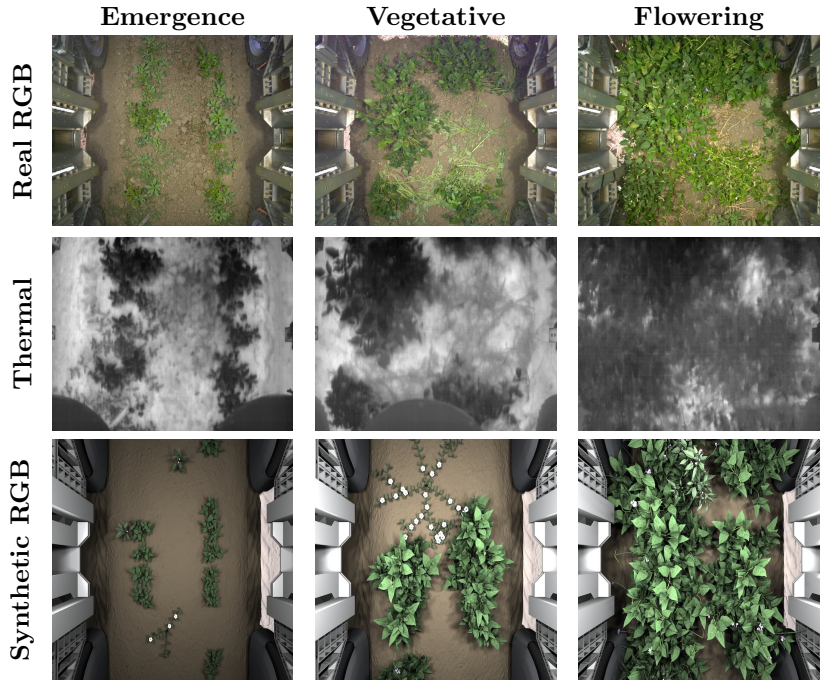
\begin{figure}[h]
    \centering
    \begin{tikzpicture}
    
        % Column titles
        \node at (-3.5,5.5) {\textbf{Emergence}};
        \node at (0,5.5) {\textbf{Vegetative}};
        \node at (3.5,5.5) {\textbf{Flowering}};
        
        % Row titles
        \node[rotate=90] at (-5.5,4) {\textbf{Real RGB}};
        \node[rotate=90] at (-5.5,1.3) {\textbf{Thermal}};
        \node[rotate=90] at (-5.5,-1.7) {\textbf{Synthetic RGB}};
        
        % Row 1 - Real RGB
        \node at (-3.5,4) {\includegraphics[width=3.2cm]{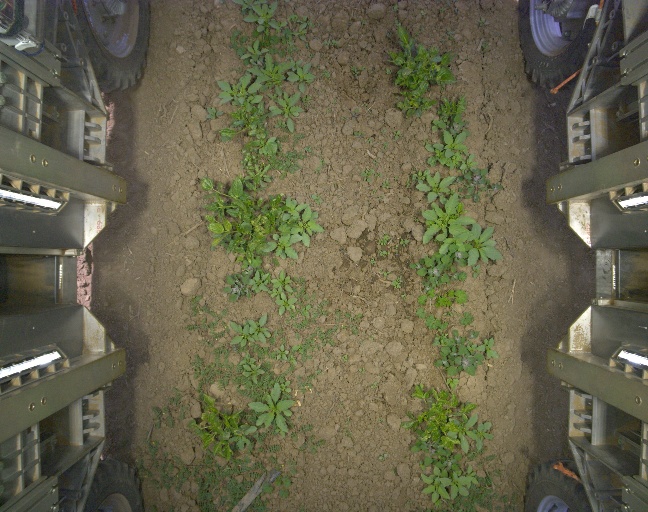}};
        \node at (0,4) {\includegraphics[width=3.2cm]{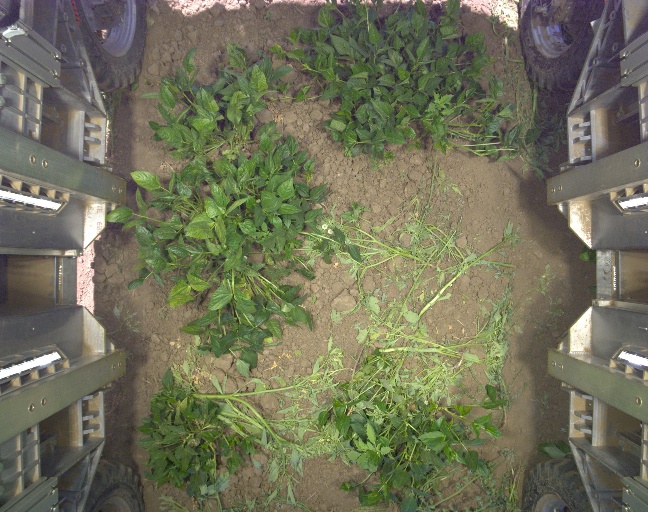}};
        \node at (3.5,4) {\includegraphics[width=3.2cm]{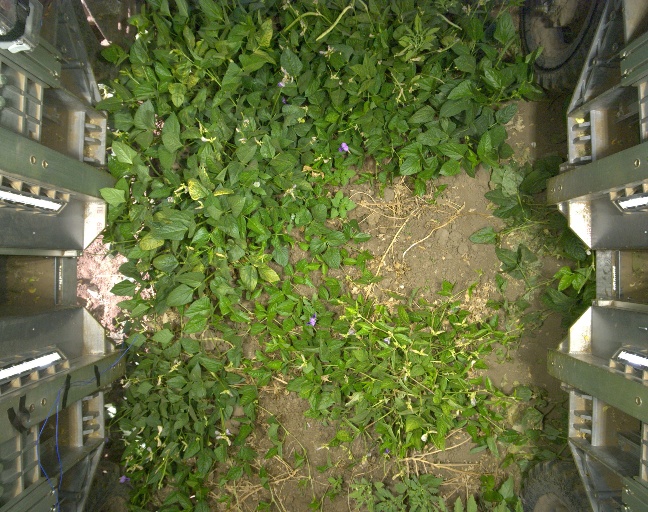}};
        
        % Row 2 - Thermal
        \node at (-3.5,1.3) {\includegraphics[width=3.2cm]{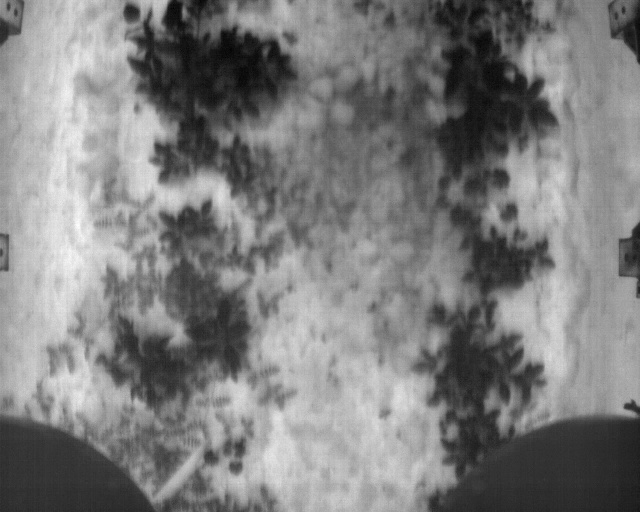}};
        \node at (0,1.3) {\includegraphics[width=3.2cm]{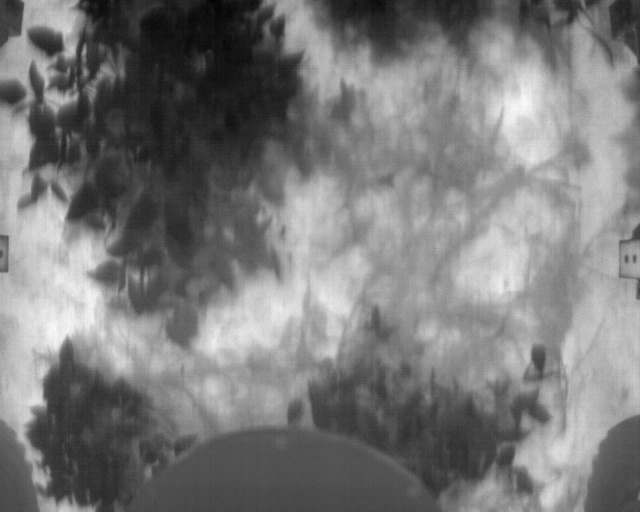}};
        \node at (3.5,1.3) {\includegraphics[width=3.2cm]{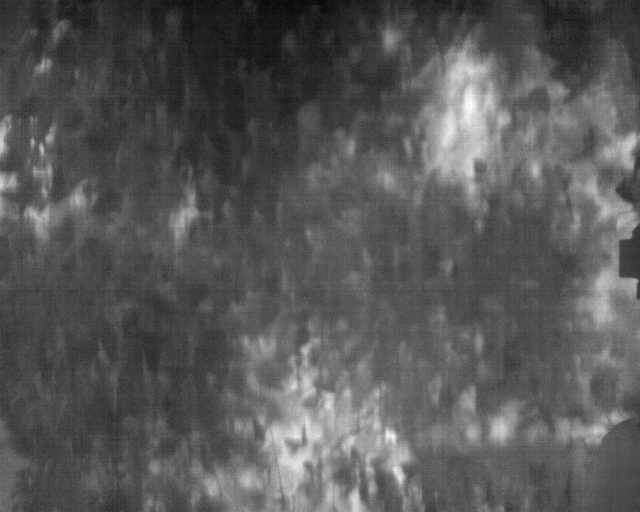}};
        
        % Row 3 - Synthetic RGB
        \node at (-3.5,-1.7) {\includegraphics[width=3.2cm]{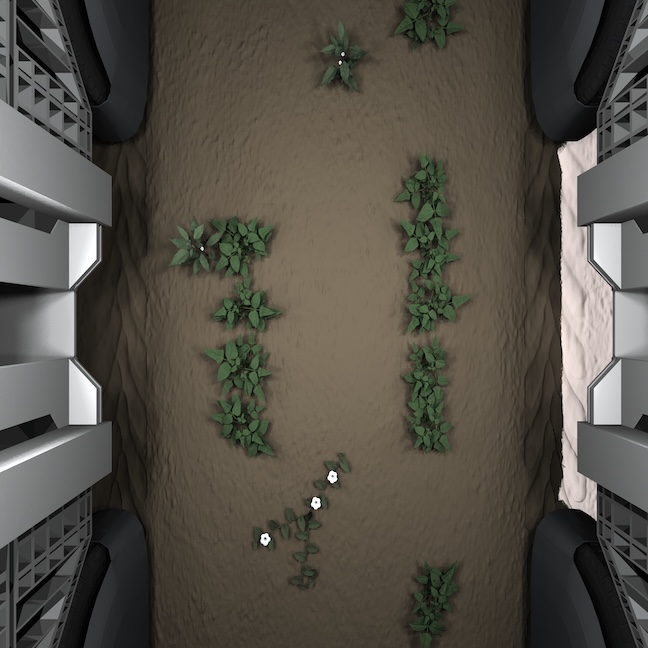}};
        \node at (0,-1.7) {\includegraphics[width=3.2cm]{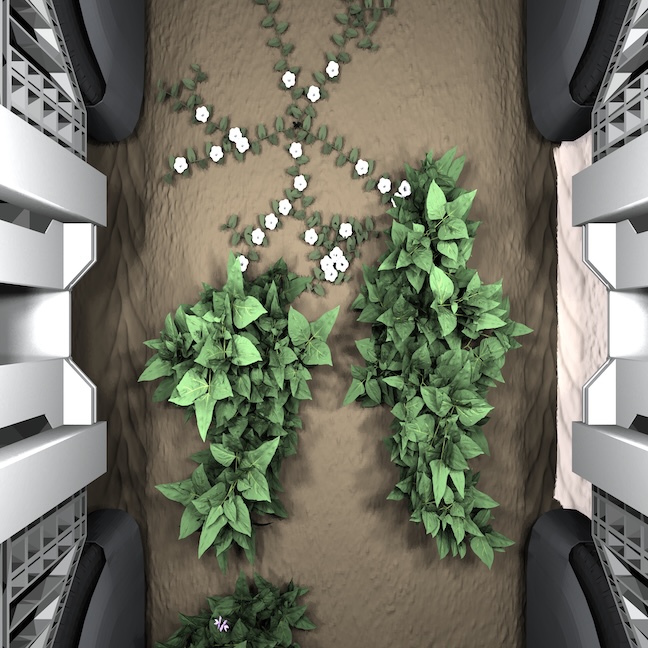}};
        \node at (3.5,-1.7) {\includegraphics[width=3.2cm]{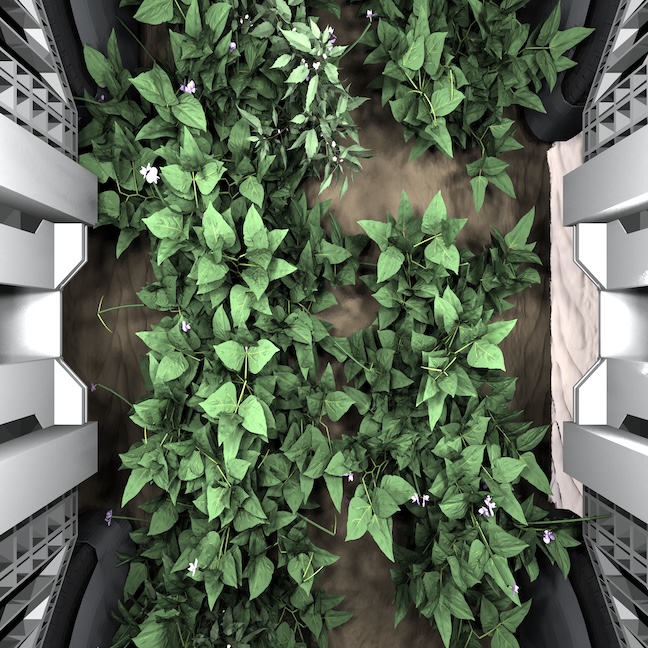}};
    
    \end{tikzpicture}
    \caption{Visual examples of real and synthetic imagery dataset from different phenological stages (emergence, vegetative, flowering) and modalities (real RGB, thermal, synthetic RGB).}
    \label{fig:phenology_grid}
\end{figure}

\begin{enumerate}
    \item Utilized synthetic images to minimize the need for manual segmentation labels, a process that is particularly time-consuming and challenging in agriculture due to the complexities of plant structures.
    \item Demonstrated that incorporating target-domain images into the training of a semantic segmentation model, originally developed with a synthetic dataset, can lead to decreased performance as model complexity increases. We highlight the trade-offs between model capacity and domain adaptation.
    \item Showed that applying strategic data sampling methods when mixing a small set of real images with a larger synthetic dataset improves generalization and reduces overfitting to the synthetic domain. 
    \item Established a RGB-to-thermal image translation pipeline that enables robust cross-modal image alignment, ensuring accurate transfer of segmentation masks between modalities.
\end{enumerate}

\subsection{Utilizing synthetic data for model training}

Recent work has demonstrated that synthetic data can provide virtually unlimited labeled examples for training models, which significantly reduces the need for costly manual annotation \cite{xu_handsoff_2023}. Additionally, it can reduce or even completely prevent incorrect labels. This strategy has been applied in domains like plant image analysis. For example, a recent leaf segmentation study showed that a model trained largely on synthetic images achieved accuracy comparable to using real images, especially when only limited real data was available \cite{hartley_domain_2024}. However, purely synthetic datasets often suffer from limited diversity and a persistent domain gap, meaning models trained on them may not generalize well to real-world inputs \cite{wang_domain_2024}. If the synthetic data lacks sufficient variety, models risk overfitting to synthetic-specific patterns and may fail to capture the complexity of real imagery. To mitigate these issues, researchers have developed techniques to make synthetic data more realistic and to bridge the synthetic-to-real gap. For instance, unsupervised domain adaptation via adversarial style transfer or recent diffusion models can render more photorealistic synthetic images while preserving labels, yielding improved performance over training on unadapted synthetic data \cite{cieslak_generating_2024, ranario_agile_2025}. Another effective approach is to mix synthetic and real data during training. By pretraining a model on abundant synthetic data and then fine-tuning it with a small set of real samples, real-world variability can be injected and significantly alleviate domain biases \cite{serrat_closing_2024}.

\subsection{Cross-modality segmentation}

Cross-modality segmentation with RGB and thermal imagery is challenging because each sensor captures different scene cues. This means that gradients and textures present in a visible image may not appear in the thermal image of the same scene (and vice versa), yielding few direct correspondences for feature matching. In practice, even if one finds common structures (e.g., object edges), aligning RGB and thermal pixels is problematic unless the sensors are carefully calibrated. Separate RGB and thermal cameras usually have different intrinsic parameters (focal length, distortion, resolution) and a relative pose offset, resulting in parallax and scale differences that break pixel-wise alignment \cite{shivakumar_pst900_2019}. Without calibration, a hot object might project to different image locations in the two modalities, compounding the difficulty of cross-domain segmentation.

One practical strategy to bridge this gap is to translate RGB images into the thermal domain using image-to-image translation, such as by using CycleGAN \cite{zhu_unpaired_2017}. By converting an RGB frame into a generated thermal image, one can create pseudo ``paired'' inputs of the same modality. For example, Wang et al. \cite{wang_cross-modality_2020} generated cross-modal paired images for person re-identification, using a generative adversarial network (GAN) to produce a thermal version of each visible image so that features can be matched instance-wise across modalities. Recent works add explicit structural constraints or advanced architectures to preserve thermal characteristics — for instance, an edge-guided multi-domain translator (EMRT) and CycleGAN-turbo to achieve more faithful thermal renditions of RGB content \cite{ma_aerialirgan_2024, lee_edge-guided_2023, parmar_one-step_2024}. On the other hand, researchers are also fusing modalities at the feature level using transformers to implicitly align RGB-thermal information. Helvig et al. \cite{helvig_caff-dino_2024} introduced a cross-attention transformer for IR-visible object detection that learns correspondences between the two modalities' feature maps. Alignment via this method may not be necessary in our case since the scenes between the RGB and thermal imagery are not vastly different.

\section{Materials and Methods}

\subsection{Overview}

First, we leverage a simulation framework to generate synthetic RGB imagery and pixel-perfect semantic masks for crops and weeds, complementing a smaller set of manually labeled real images collected in a cowpea breeding trial using a ground-based rover equipped with visible and thermal cameras. To bridge the synthetic-to-real domain gap, we investigate multiple strategies for integrating real images into the synthetic training set, including methods called: direct injection, balanced sampling, and fine-tuning. A semantic segmentation model is trained using various encoder-decoder configurations and optimized with multi-class Dice loss. Predicted RGB segmentation masks are then transferred to corresponding thermal images using a GAN-based template matching pipeline, which aligns RGB and thermal views. This enables extraction of plant temperature data while excluding weeds and background.

\subsection{Dataset}

\begin{table}[t]
\centering
\caption{Camera specifications for RGB and thermal imaging.}
    \begin{tabular}{|l|c|c|}
        \hline
        \textbf{Parameter} & \textbf{Basler acA2500-20gc (RGB)} & \textbf{FLIR Boson 640 (Thermal)} \\
        \hline
        Resolution (pixels) & 2592 × 2048 & 640 × 512 \\
        Pixel Size ($\mu$m) & 4.8 × 4.8 & 12 x 12 \\
        Focal Length (mm) & Lens-dependent & 8.7 \\
        Field of View (HFOV) & Lens-dependent & 50$^\circ$ \\
        Shutter Type & Global shutter & Rolling (thermal) \\
        Spectral Band & Visible (RGB) & 8–14 $\mu$m (LWIR) \\
        Principal Point ($c_x$, $c_y$) & Image center (approx.) & Image center (approx.) \\
        Frame Rate (Hz) & Up to 21 & Up to 60 \\
        \hline
    \end{tabular}
\label{tab:camera_specs}
\end{table}

\begin{figure}
    \centering
    \includegraphics[width=0.8\linewidth]{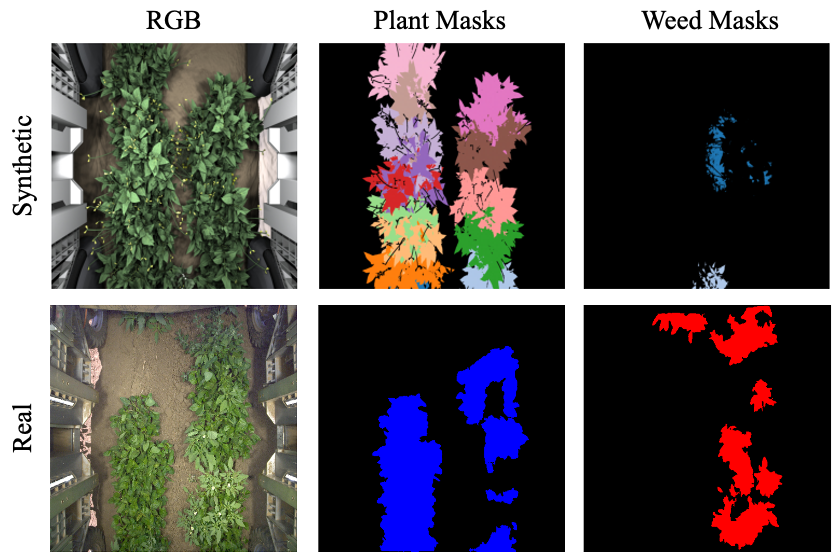}
    \caption{\textit{Top row}: Synthetic instance segmentation masks generated from Helios \cite{bailey_helios_2019,lei_simulation_2024}. For semantic segmentation, each instance is collated into a single mask per class (plant and weed). \textit{Bottom row}: Example real, manually labeled semantic segmentation masks for plant (blue) and weed (red) classes.}
    \label{fig:syn_real_masks}
\end{figure}

\subsubsection{Real images}
Images were collected within a cowpea crop ({\em Vigna unguiculata}) \cite{huynh_multi-parent_2018} breeding trial using a ground-based rover (T4-146, Mineral, Mountain View, CA, USA) across multiple time points. The imaging system on the rover includes both visible and thermal spectrum sensors. The RGB images were acquired using a Basler acA2500-20gc (Basler AG, Ahrensburg, Germany) color area scan camera. This camera captures images at a resolution of 2592x1944 pixels. The rover utilized a closed canopy to block ambient sunlight, and illuminated plants under the canopy using a XLamp XHP70.2 6500K RGB LED lights by Cree (Cree LED, Durham, NC, USA), ensuring consistent lighting conditions during image capture. The long-wave infrared thermal imagery was collected using a FLIR Boson 640 (Teledyne FLIR,
Wilsonville, Oregon) thermal camera, configured with an 8.7 mm lens with a 50\degree horizontal field of view (HFOV). This sensor captures images with a resolution of 640x512 pixels. More details for each camera are summarized in Table \ref{tab:camera_specs} and a sample image is shown in Figure \ref{fig:syn_real_masks}.

sImage data were collected at eight different time points, each spaced approximately five days apart. At each time point, image acquisition was conducted at the same time of day, yielding an average of approximately \(11,000\) images per camera. For this study, a representative subset of 35 images spanning multiple phenological stages was manually annotated. To maximize the relevance of the labeled data, images containing a substantial presence of weeds, defined as any non-cowpea plant within the frame, were prioritized during selection. Refer to Figure \ref{fig:num_pixels_per_class} for the number of pixels per class that is represented within the batch of real images.

\subsubsection{Synthetic images}

A synthetic RGB imagery dataset was generated using Helios \cite{bailey_helios_2019,lei_simulation_2024}, an open source 3D plant and environment biophysical modeling framework. The current version of Helios can be downloaded at \url{https://www.github.com/PlantSimulationLab/Helios}.

Helios uses a unique, radiation modeling approach to generate synthetic images and corresponding segmentation masks. Cowpea and weed model plants were generated at growth stages matching the field imagery using the ``plant architecture" plug-in of Helios, which is a parametric procedural plant architectural model. Weed species models included in the synthetic images were cheeseweed ({\em Malva neglecta}), puncturevine ({\em Tribulus terrestris}), bindweed ({\em Convolvulus arvensis}), and ground cherry ({\em Physalis philadelphica}), all of which are openly available in the Helios plant library. Each architectural parameter in the models can be specified based on a random distribution in order to create variability in the images.

The camera model requires specification of the spectral reflectivity and transmissivity of all surfaces in the simulated scene, the spectral intensity of the light source, and the camera spectral response and intrinsic parameters. Various measured surface reflectivity and transmissivity spectra are available in the Helios spectral library, which can be used to introduce variability in the image set. The light source and camera properties were specified based on the manufacturer data sheets. Segmentation masks were generated by assigning a unique integer ID to each element comprising each plant or weed in the image, which is then used by Helios to automatically generate pixel ID maps for each image, as shown in Figure \ref{fig:syn_real_masks}. A more detailed description and evaluation of the synthetic imagery generation methodology can be found in \cite{lei_simulation_2024}.

Typically, synthetic images are treated as training data based on their capability to generate a virtually limitless amount of labeled images. Additionally, plant modeling parameters can be tweaked to close the domain gap, in exchange for resources and compute time, to allow for improved model performance. The synthetic images were curated to closely match the real images, as they also include rover parts, as seen in Figure \ref{fig:syn_real_masks}. A timeseries instance segmentation dataset of 1128 images was created with the following classes: plant and weed. 
% There are in total 1128 instances of plants and 1027 instances of weeds, with a total of 1128 images.

\begin{figure}[t]
    \centering
    \includegraphics[width=0.8\linewidth]{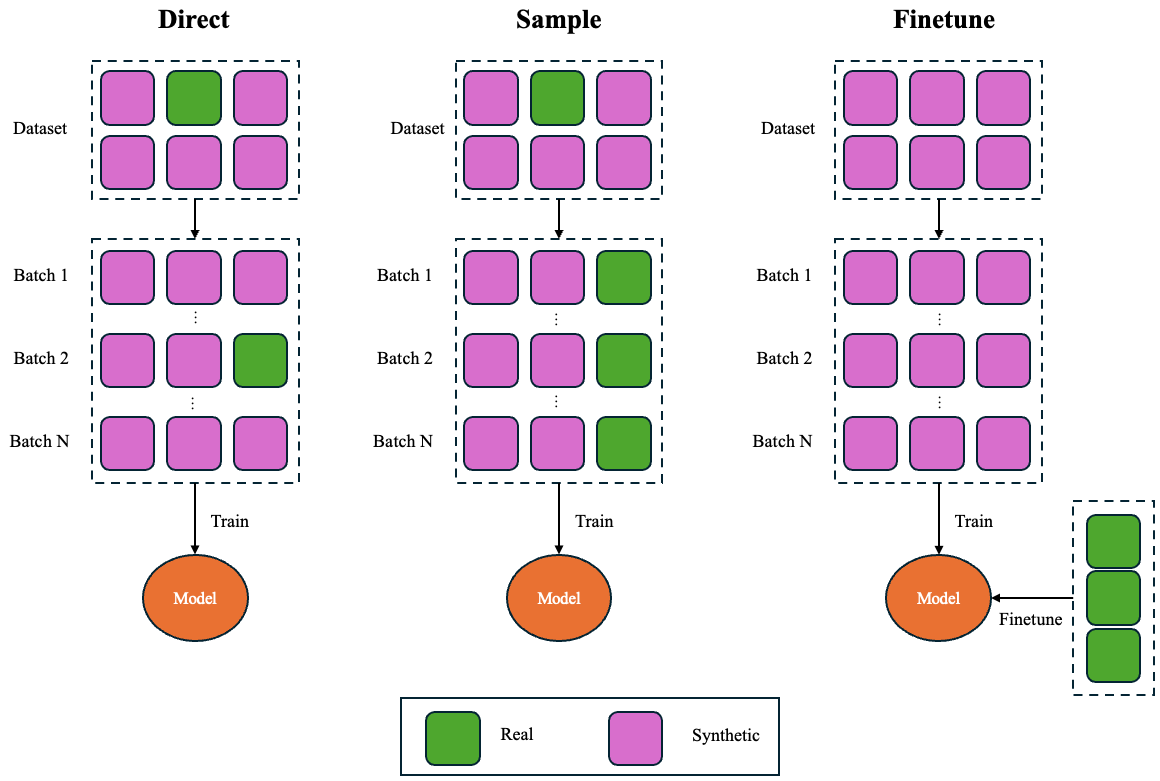}
    \caption{Sampling methods when injecting real images into a synthetic dataset. \textit{Direct} injection, where real images are added to the synthetic dataset without sampling constraints. \textit{Balanced} sampling, where each training batch includes at least one real image. \textit{Fine-tuning}, where the model pretrained on synthetic data is subsequently fine-tuned using the selected real images.}
    \label{fig:sampling}
\end{figure}

\subsection{Semantic segmentation}

\subsubsection{Injecting real images into a synthetic dataset}

Training with the synthetic images enables efficient model development and virtually unlimited dataset scaling, since simulation frameworks can generate thousands of labeled samples on demand. But, at the same time, model performance remains tightly coupled to the quality and diversity of its training data \cite{montesinos_lopez_overfitting_2022, xu_splitting_2018}. Although our synthetic images have been curated to resemble the real images, models trained exclusively on them tend to under-perform on actual imagery due to a persistent “synthetic-to-real” domain gap \cite{man_review_2022}. Nevertheless, synthetic data still encapsulates structural and contextual features that, when paired with appropriate augmentations, can drive strong learning outcomes. For instance, Cakić et al. \cite{cakic_evaluating_2025} showed that hybrid datasets with roughly 30–50\% synthetic data content delivered higher detection accuracy and cross-domain robustness than purely real or purely synthetic sets.

Our training set consisted of 1,128 synthetic images, while the validation and test sets each included 10 real images with manual annotations. Although Helios can produce an essentially limitless stream of synthetic data, our experiments (see Figure \ref{fig:scaling_syn}) show that generating beyond approximately 1,000 images offers no measurable benefit for this task. Moreover, because the synthetic samples were designed to capture the same key visual characteristics present in the real images, the feature distributions between the two domains remain closely aligned. 

An additional 15 real images were reserved to assess how model performance evolved as real data was progressively integrated into the synthetic training pool. We investigated three strategies for incorporating real data, as seen in Figure~\ref{fig:sampling}: (1) \textit{direct} injection, where real images are added to the synthetic dataset without sampling constraints; (2) \textit{balanced} sampling, where each training batch includes at least one real image; and (3) \textit{fine-tuning}, where the model pretrained on synthetic data is subsequently fine-tuned using the selected real images. For all sampling strategies, real images were augmented using a diverse set of transformations, including horizontal flipping, shift-scale cropping, Gaussian noise, perspective distortion, and various color perturbations, to enhance variability and reduce overfitting given the limited size of the dataset. The Albumentations library \cite{buslaev_albumentations_2020} was used to apply the augmentations to the training set and exact augmentation values can be found in Table~\ref{tab:augmentations}.

\subsubsection{Model architecture - decoders}

To determine which decoder design best complements the encoder architecture and target application, several representative decoder structures were evaluated. Each option offered distinct trade-offs between complexity, multi-scale feature fusion, and semantic granularity.

UNet++ \cite{zhou_unet_2018} is an advanced encoder-decoder architecture for semantic segmentation that introduces nested and dense skip connections to bridge the semantic gap between encoder and decoder feature maps. Unlike the original U-Net, which directly forwards encoder features to the decoder, UNet++ enriches these features through intermediate convolutional blocks, making them semantically closer to their corresponding decoder layers. This nested design simplifies optimization and enhances fine-grained detail recovery. Additionally, UNet++ integrates deep supervision, allowing for flexible model pruning and improving performance across multiple segmentation tasks, especially in the medical imaging domain. However, this nested design comes with the cost of increased computational overhead and memory usage, potentially limiting resource constrained hardware \cite{yin_unet--_2024}. 

Feature Pyramid Network (FPN) \cite{lin_feature_2017} is designed to efficiently handle multi-scale object recognition by leveraging the inherent pyramidal hierarchy of convolutional neural networks (CNNs). It combines low-resolution, semantically strong features with high-resolution, spatially precise features through a top-down pathway with lateral connections. This design enables FPN to produce feature maps with rich semantics at all scales, significantly enhancing performance in object detection and segmentation tasks. Importantly, FPN achieves this without the computational burden of traditional image pyramids, making it both effective and efficient. The trade-off is that as the number of parameters increase, the architectural complexity can make model training difficult in custom pipelines \cite{chen_info-fpn_2023}.

SegFormer \cite{xie_segformer_2021} is a transformer-based semantic segmentation framework that balances efficiency, accuracy, and robustness. It introduces a hierarchical Transformer encoder \cite{vaswani_attention_2023} without positional encodings, which produces multi-scale features that generalize well across varying input resolutions. SegFormer also fuses features from multiple stages, combining local and global context effectively. This simple yet powerful architecture avoids heavy computation while achieving state-of-the-art performance on several benchmarks, making it ideal for both real-time and high-accuracy applications. The limitation is that scaling model size requires careful hyperparameter fine-tuning and dropout \cite{bai_dynamically_2021}.

\subsubsection{Model architecture - encoders}

EfficientNet \cite{tan_efficientnet_2020} is a family of CNNs designed for computational efficiency while maintaining accuracy. The key innovation is a compound scaling method that uniformly scales the network's depth, width, and input resolution using fixed coefficients. The compound method scales depth, width, and resolution together according to user-defined coefficients, ensuring balanced model scaling with image resolution under a fixed computational budget. This stands in contrast to traditional approaches that scale only one dimension of the model, often leading to diminishing returns. EfficientNet models are built upon a baseline architecture called EfficientNet-b0, which was discovered using neural architecture search. Larger models (e.g., b1–b8) are then derived by scaling b0 using the compound method, offering a principled way to trade off between accuracy and computational cost. We test the b0, b5 and b8 variants as our encoders to evaluate how the number of trainable parameters influence model performance and domain adaptation.

\subsubsection{Loss function}

Before evaluation, the model is trained using a \textit{multi-class Dice loss} \cite{milletari_v-net_2016}. This loss function directly optimizes spatial overlap between predicted and ground truth regions across all classes, making it especially suitable for imbalanced segmentation tasks. For each training batch, the network outputs unnormalized logits, which are passed directly to the Dice loss. The multi-class Dice loss is computed by comparing each class prediction channel against the corresponding ground truth binary mask and averaging the per-class Dice scores. The loss is defined as:

\begin{equation}
\mathcal{L}_{\text{Dice}} = 1 - \frac{1}{C} \sum_{c=1}^{C} \frac{2 \sum_{i} p_{i,c} \cdot g_{i,c} + \epsilon}{\sum_{i} p_{i,c}^2 + \sum_{i} g_{i,c}^2 + \epsilon}
\end{equation}

where:
\begin{itemize}
    \item $C$ is the number of classes,
    \item $p_{i,c}$ is the predicted softmax probability for pixel $i$ belonging to class $c$,
    \item $g_{i,c}$ is the corresponding one-hot ground truth label (0 or 1),
    \item $\epsilon$ is a small constant to prevent division by zero.
\end{itemize}

This formulation encourages the model to maximize overlap between predicted and ground truth masks for each class. Compared to cross-entropy, Dice loss is more robust to class imbalance, as it directly penalizes both false positives and false negatives in its overlap computation.

\subsubsection{Evaluation criteria and metrics}

To quantitatively evaluate segmentation performance, we use three standard metrics: Intersection over Union (IoU) \cite{everingham_pascal_2010}, Dice coefficient, and Pixel Accuracy \cite{cordts_cityscapes_2016}. These metrics are calculated \textit{per pixel} across the entire dataset, making them well-suited for evaluating dense prediction tasks like semantic segmentation. Each pixel is treated as an independent classification decision, assigned to a single class label.

For a multi-class segmentation problem with $C$ classes, the confusion statistics are computed for each class $c \in \{1, \dots, C\}$ by comparing predicted pixel labels $\hat{y}_{ij}$ and ground truth labels $y_{ij}$ over all pixel locations $(i,j)$ in an image:
\begin{itemize}
    \item \textbf{True Positives (TP):} Pixels correctly predicted as class $c$.
    \item \textbf{False Positives (FP):} Pixels incorrectly predicted as class $c$ (they belong to another class in ground truth).
    \item \textbf{False Negatives (FN):} Pixels belonging to class $c$ in the ground truth but predicted as another class.
    \item \textbf{True Negatives (TN):} Pixels correctly predicted as not belonging to class $c$.
\end{itemize}

IoU measures the agreement between the predicted and ground truth regions, penalizing both oversegmentation and undersegmentation. For class $c$, it is computed as:
\begin{equation}
\text{IoU}_c = \frac{\text{TP}_c}{\text{TP}_c + \text{FP}_c + \text{FN}_c + \epsilon}
\end{equation}
where $\epsilon$ is a small constant (e.g., $1e^{-7}$) to ensure numerical stability. A high IoU indicates precise localization and accurate region coverage.

The Dice coefficient emphasizes overlap and is particularly sensitive to class imbalance, making it a complementary metric to IoU. It is defined as:
\begin{equation}
\text{Dice}_c = \frac{2 \cdot \text{TP}_c}{2 \cdot \text{TP}_c + \text{FP}_c + \text{FN}_c + \epsilon}
\end{equation}
This metric is commonly used in medical imaging and plant phenotyping due to its robustness to small region errors.

Pixel Accuracy evaluates the overall proportion of pixels correctly classified (either as class $c$ or not):
\begin{equation}
\text{PixelAccuracy}_c = \frac{\text{TP}_c + \text{TN}_c}{\text{TP}_c + \text{FP}_c + \text{FN}_c + \text{TN}_c + \epsilon}
\end{equation}
While intuitive, pixel accuracy can be misleading in imbalanced datasets dominated by background pixels.

These metrics help account for class imbalance and provide a balanced view of model performance. In addition to aggregate scores, we also report per-class performance for interpretability and diagnostic purposes. For example, in our case, metrics are computed for background, weed, and plant classes. To completely evaluate model performance, we make the following comparisons: to train using only synthetic data, only real data, and using both synthetic and real data.

\begin{figure}[t]
    \centering
    \includegraphics[width=0.8\linewidth]{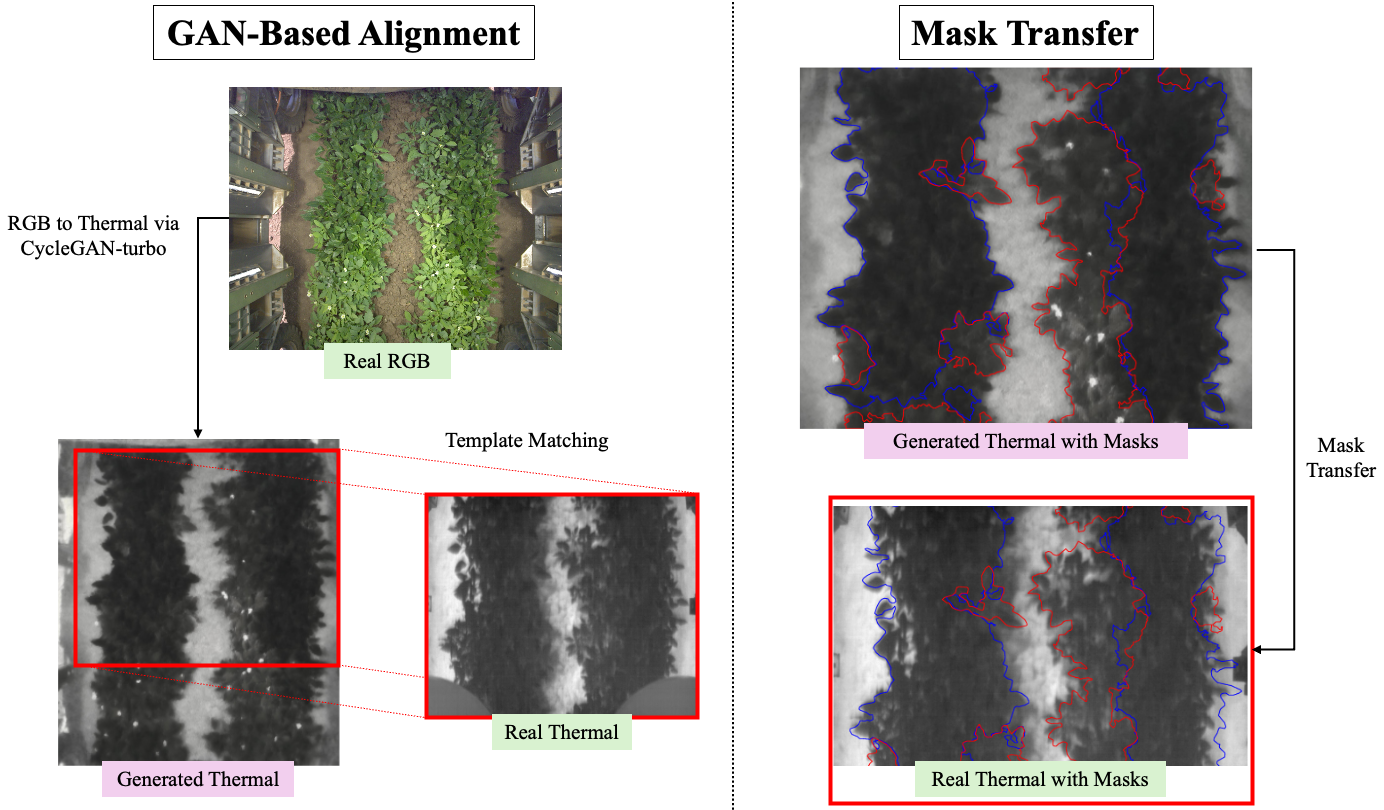}
    \caption{Overview of our GAN-based alignment and mask transfer method. We first translate the RGB image into the thermal domain. This enables template matching between the generated and real thermal images. Once the matched region is located, the segmentation mask from the RGB image can be mapped onto the target thermal image.}
        \label{fig:alignment_transfer}
\end{figure}

\subsection{GAN-based alignment}

After performing segmentation on the RGB images, we transfer the resulting masks to their corresponding thermal images. To accomplish this, we first identify the matched regions between the RGB and thermal views. We translate the RGB images into the thermal domain using CycleGAN-turbo \cite{parmar_one-step_2024}, enabling direct template matching between the generated and real thermal images. Once the matched region is located, the segmentation mask from the RGB image is mapped onto the thermal image. Notably, this approach does not require explicit geometric alignment between the RGB and thermal images, as the template matching operates effectively without camera calibration.

To align generated thermal (RGB-translated) and real thermal images, we apply a multi-scale template matching algorithm. Since the RGB and thermal cameras have different intrinsic parameters and have a relative offset, we search for corresponding RGB-translated images within a defined time frame. Then, each candidate RGB-translated image is searched over multiple scales (from 0.1× to 1.0×), and normalized cross-correlation is used to identify the region of highest similarity (more details in Figure \ref{fig:template_matching}). The scale and location with the best match are recorded and used to compute the bounding box coordinates of the matched region in the original RGB image. This approach enables effective alignment of images across modalities without requiring camera calibration or geometric rectification. This approach is summarized in Figure \ref{fig:method} and Figure \ref{fig:alignment_transfer}. 

\begin{figure}[h]
    \centering
    \includegraphics[width=0.7\linewidth]{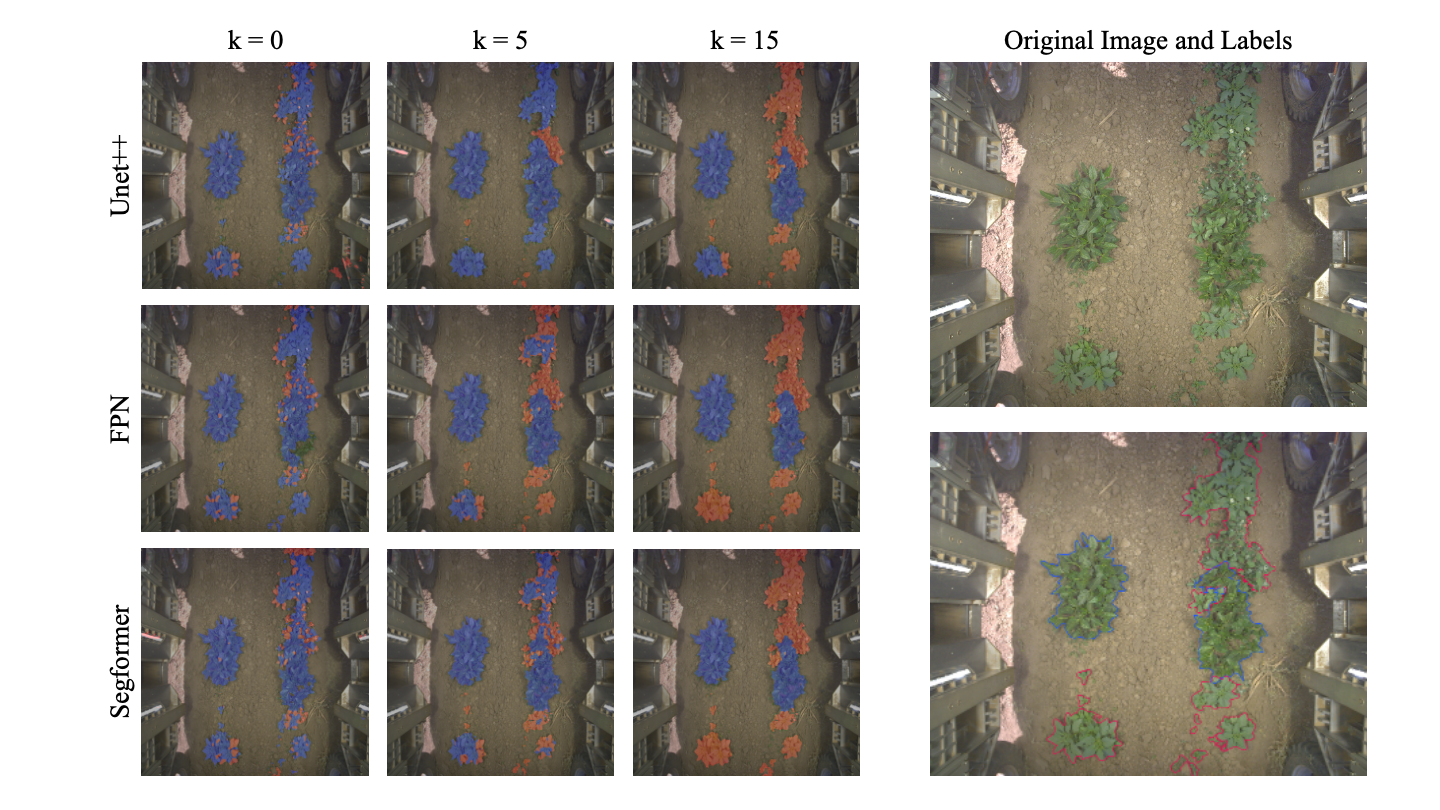}
    \includegraphics[width=0.7\linewidth]{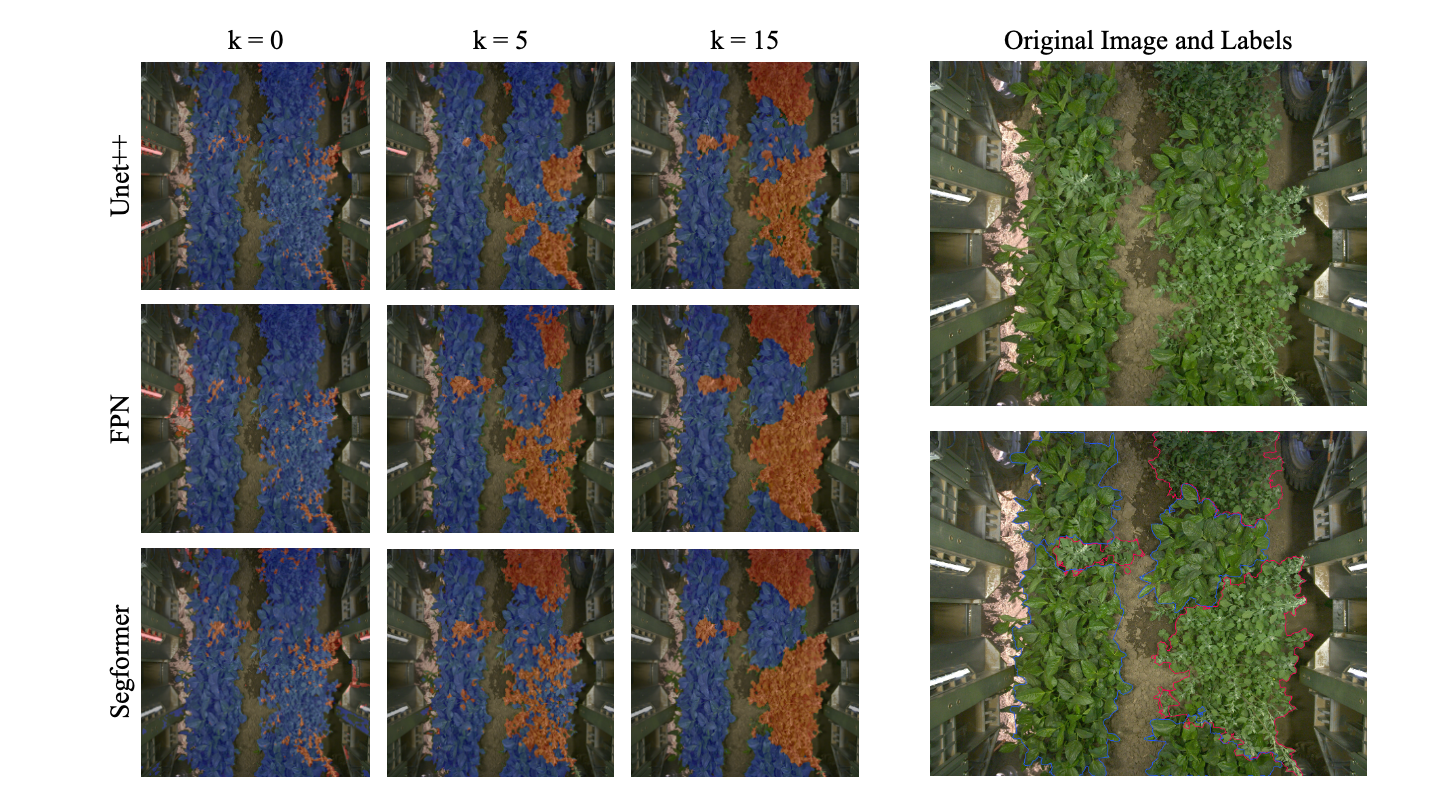}
    \caption{Sample predictions across different model architectures (each are 31M variants). Blue masks are plants and red masks are weeds. Each column represents the result with $k$ number of target images included in the training set.}
    \label{fig:sample_predictions}
\end{figure}

\section{Results}

\subsection{Data scaling model performance with real data}

\begin{table}[t]
\centering
\scriptsize
    \caption{Comparison of segmentation models using different encoders across IoU, Dice, and Pixel Accuracy metrics using a synthetic only dataset. Bold numbers are the highest mean performance values across all models for the respective metric. Each b0 variant has 6M parameters, b5 has 31M parameters, and b8 has 86M parameters.}
    \begin{adjustbox}{width=\textwidth}
        \begin{tabular}{llcccccccccccc}
        \toprule
            \multirow{2}{*}{Decoder} & \multirow{2}{*}{Encoder} & \multicolumn{4}{c}{IoU (\%)} & \multicolumn{4}{c}{Dice (\%)} & \multicolumn{4}{c}{Pixel Accuracy (\%)} \\
            \cmidrule(lr){3-6} \cmidrule(lr){7-10} \cmidrule(lr){11-14}
            & & Other & Weed & Plant & \textit{Mean} & Other & Weed & Plant & \textit{Mean} & Other & Weed & Plant & \textit{Mean} \\
        \midrule
            \multirow{3}{*}{UNet++} 
            & Efficientnet-b0 & 0.949 & 0.095 & 0.328 & 0.458 & 0.974 & 0.181 & 0.475 & 0.537 & 0.954 & 0.921 & 0.951 & 0.941 \\
            & Efficientnet-b5 & 0.959 & 0.123 & 0.524 & 0.535 & 0.979 & 0.208 & 0.659 & 0.615 & 0.964 & 0.944 & 0.953 & 0.954 \\
            & Efficientnet-b8 & 0.965 & 0.106 & 0.543 & 0.538 & 0.982 & 0.163 & 0.669 & 0.611 & 0.969 & 0.944 & 0.947 & 0.955 \\
        \midrule
            \multirow{3}{*}{FPN} 
            & Efficientnet-b0 & 0.963 & 0.171 & 0.376 & 0.503 & 0.981 & 0.272 & 0.527 & 0.594 & 0.968 & 0.938 & 0.950 & 0.952 \\
            & Efficientnet-b5 & 0.969 & 0.176 & 0.550 & \textbf{0.565} & 0.984 & 0.278 & 0.673 & \textbf{0.645} & 0.973 & 0.950 & 0.952 & 0.958 \\
            & Efficientnet-b8 & 0.962 & 0.150 & 0.492 & 0.534 & 0.980 & 0.242 & 0.620 & 0.614 & 0.966 & 0.944 & 0.955 & 0.955 \\
        \midrule
            \multirow{3}{*}{Segformer} 
            & Efficientnet-b0 & 0.966 & 0.057 & 0.528 & 0.517 & 0.982 & 0.097 & 0.654 & 0.578 & 0.971 & 0.943 & 0.945 & 0.953 \\
            & Efficientnet-b5 & 0.964 & 0.167 & 0.553 & 0.561 & 0.981 & 0.265 & 0.674 & 0.640 & 0.969 & 0.946 & 0.951 & 0.955 \\
            & Efficientnet-b8 & 0.964 & 0.298 & 0.393 & 0.552 & 0.982 & 0.436 & 0.509 & 0.642 & 0.969 & 0.950 & 0.961 & \textbf{0.960} \\
        \bottomrule
        \end{tabular}
    \end{adjustbox}
\label{tab:syn_results}
\end{table}

\begin{figure}[h]
    \centering

    \begin{subfigure}[c]{0.45\linewidth}
        \includegraphics[width=\linewidth]{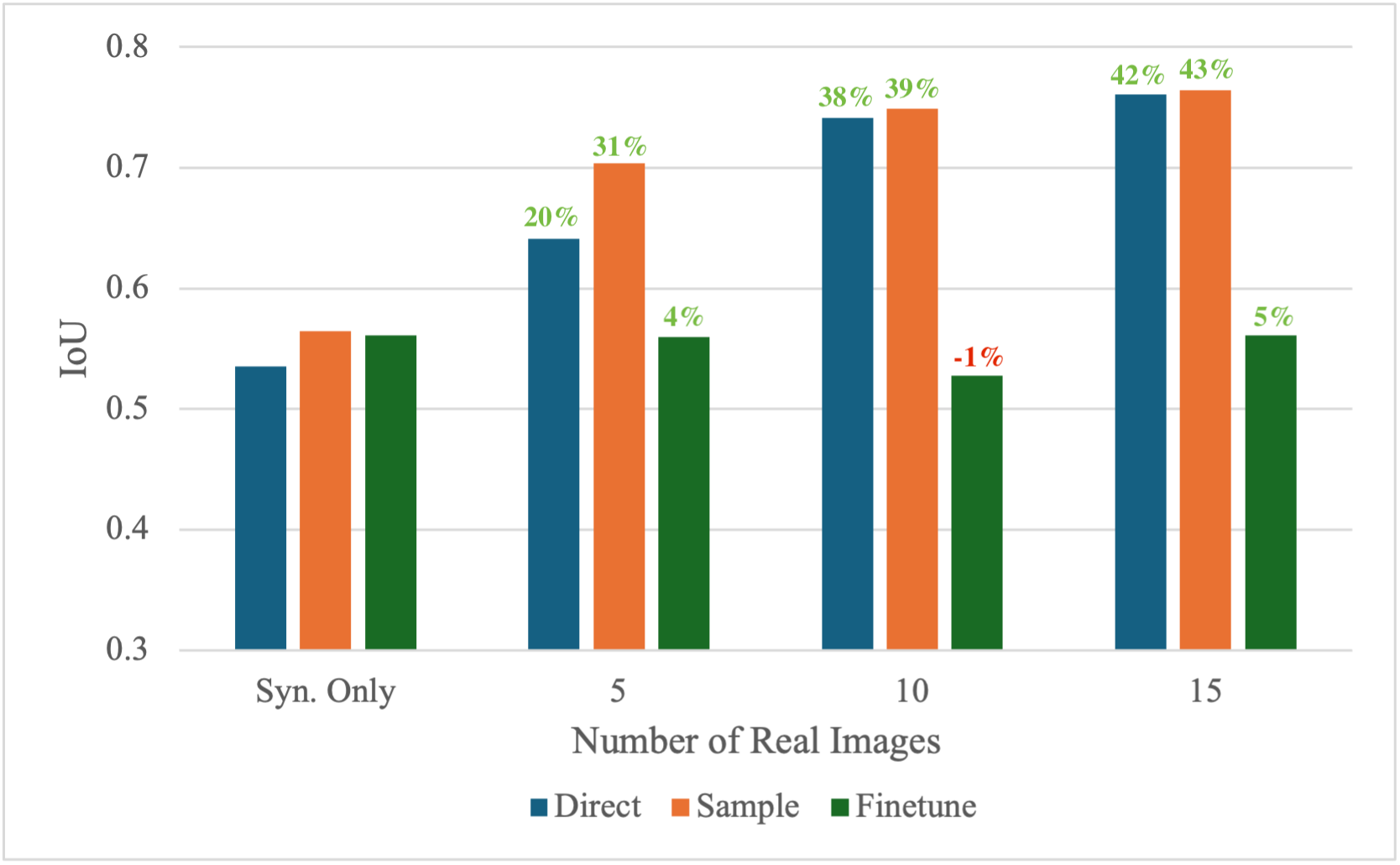}
        \caption{Using the UNet++ (31M) model, a comparison between a synthetic-only dataset and a synthetic plus $k$ real dataset using mean IoU scores for each real data injection method: direct, sample, and finetune.}
        \label{fig:syn_vs_real}
    \end{subfigure}
    \hfill
    \begin{subfigure}[c]{0.45\linewidth}
        \includegraphics[width=\linewidth]{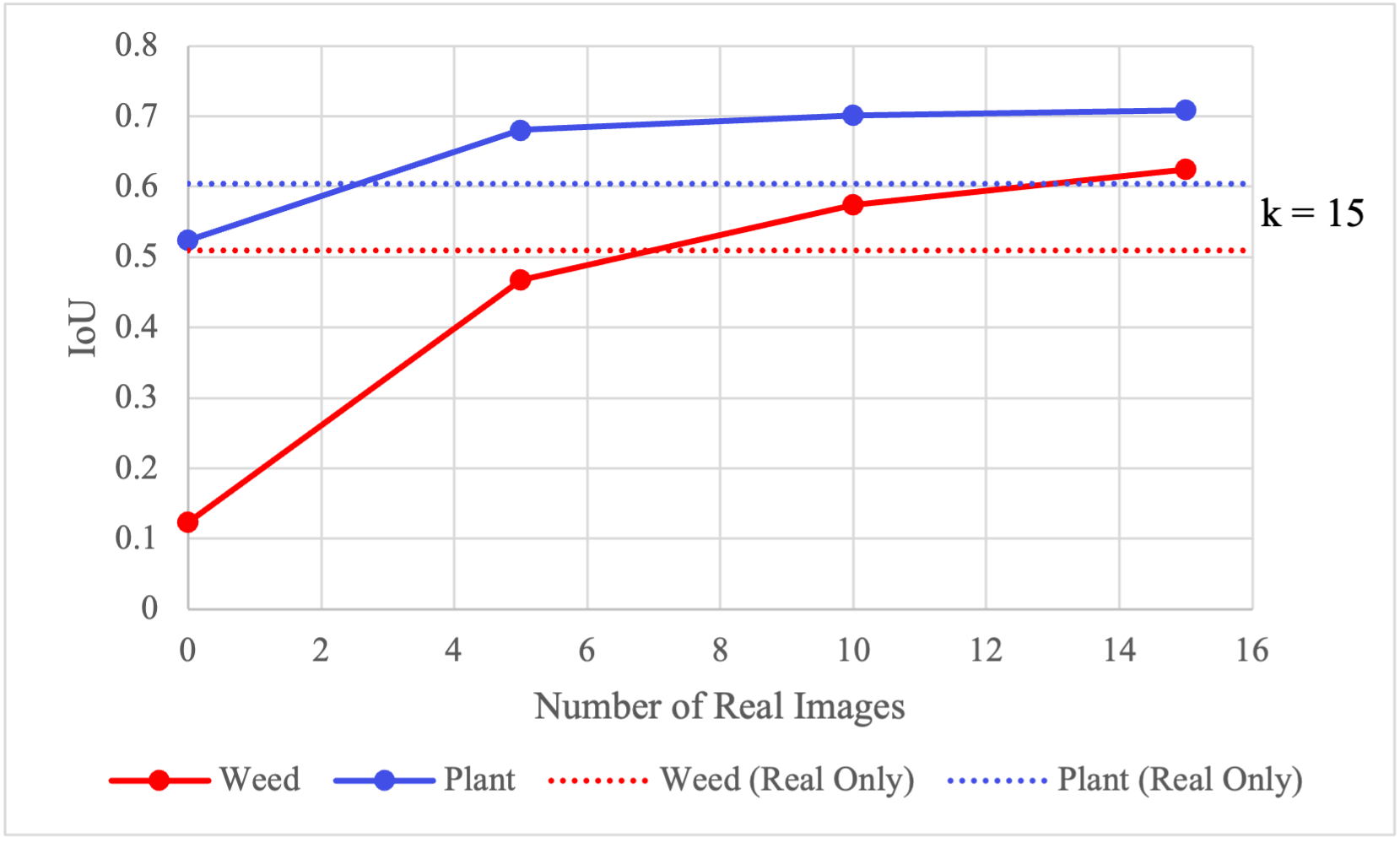}
        \caption{Using the UNet++ (31M) model, weed and plant class IoU performance for $k$ real images included into the synthetic dataset using the balanced sampling method. Horizontal lines mark test results using only real data for training.}
        \label{fig:class_k}
    \end{subfigure}

    \vspace{0.5em}

    \begin{subfigure}[c]{0.45\linewidth}
        \includegraphics[width=\linewidth]{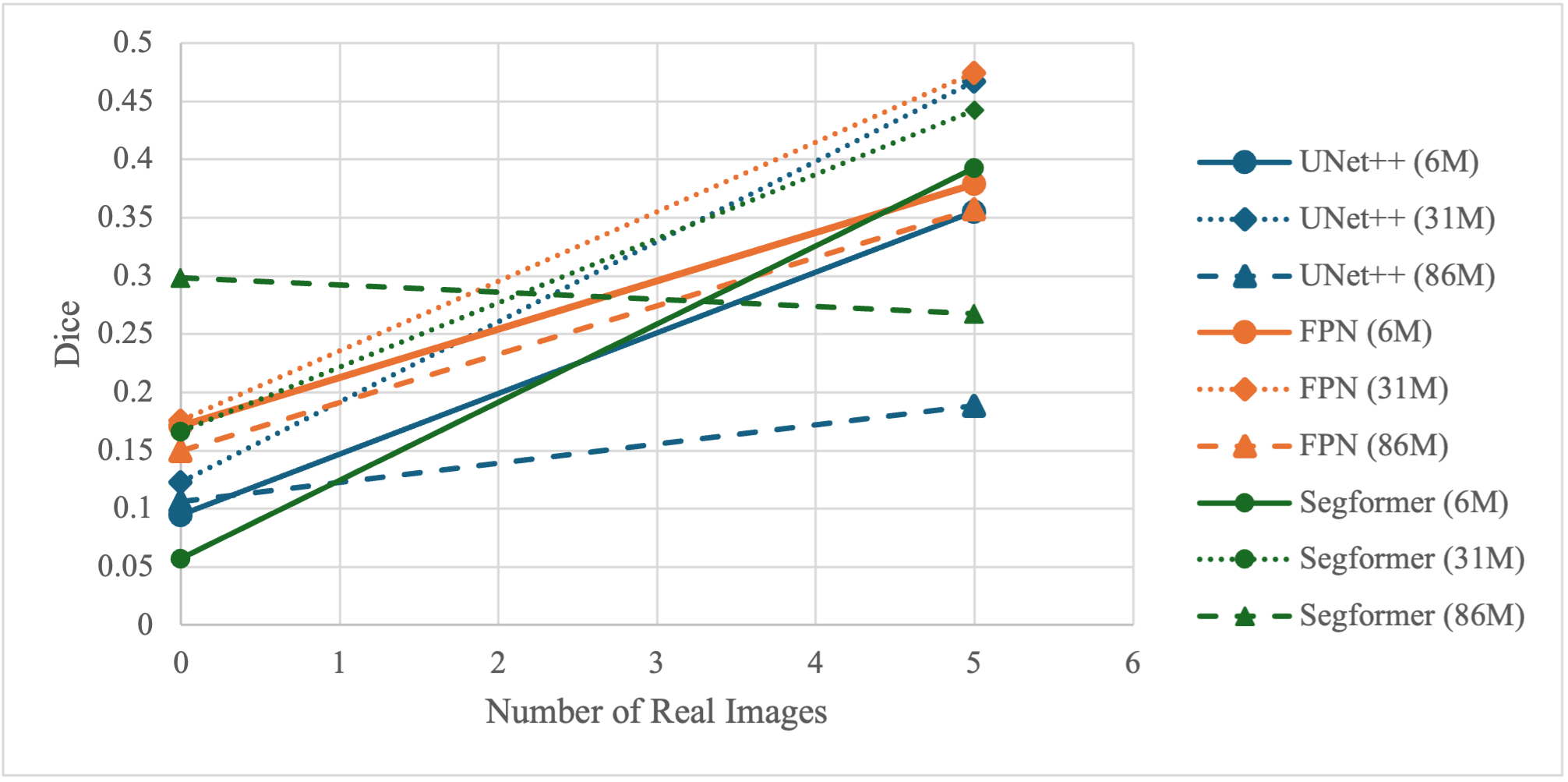}
        \caption{Significant increase in Weed Dice scores with only five real data points. Results are shown across different model scales and parameter sizes.}
        \label{fig:model_k_5}
    \end{subfigure}
    \hfill
    \begin{subfigure}[c]{0.45\linewidth}
        \includegraphics[width=\linewidth]{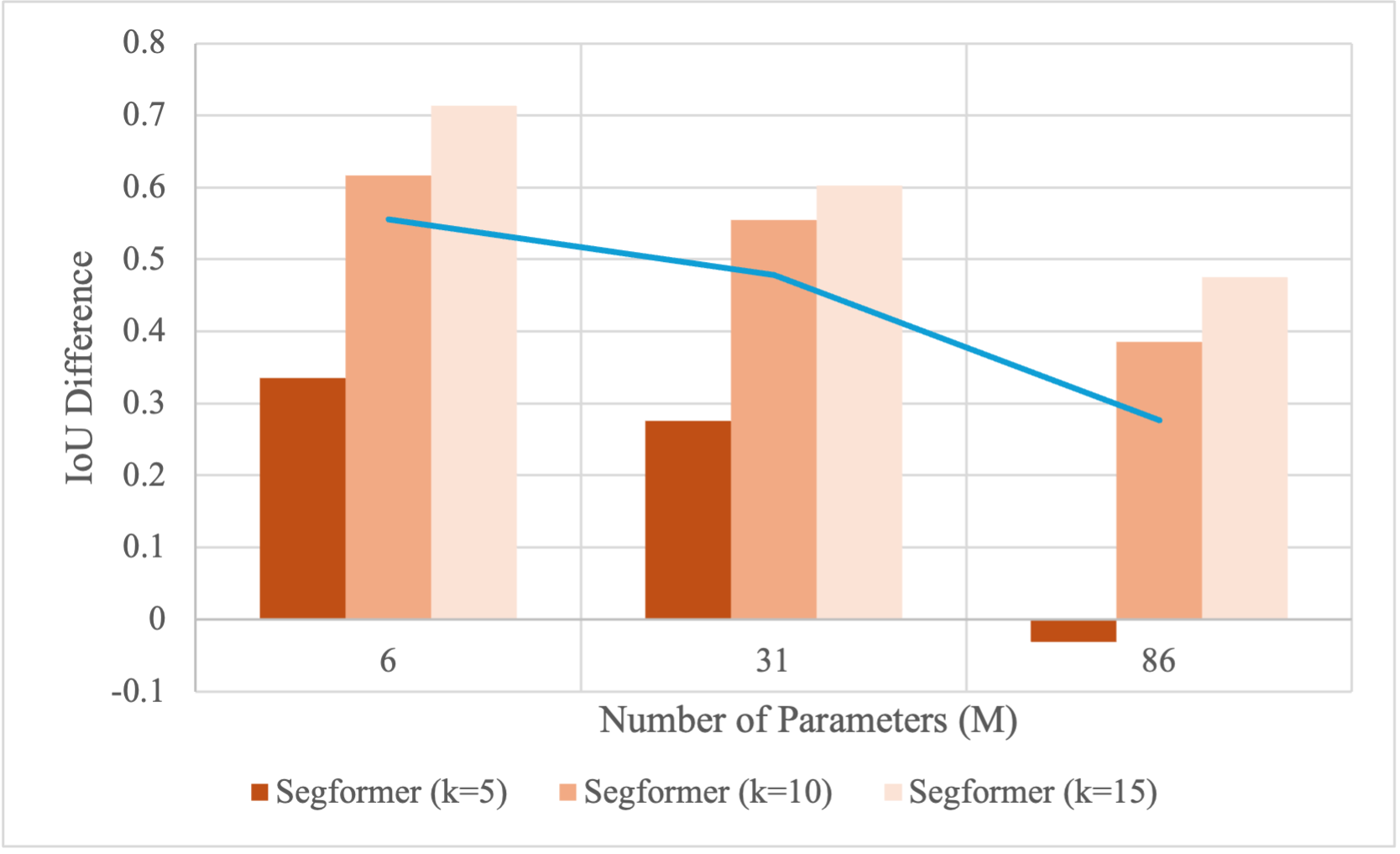}
        \caption{Impact of model parameter size on performance. Shown as IoU improvement between sample-based training (k=5,10,15) and baseline (k=0).}
        \label{fig:segformer_model_size}
    \end{subfigure}

    \caption{Comparisons across real data inclusion strategies, class-specific trends, and model scale. Total real images used is $k=15$.}
    \label{fig:full_comparison_grid}
\end{figure}

\begin{figure}[h]
    \centering
    \includegraphics[width=0.7\linewidth]{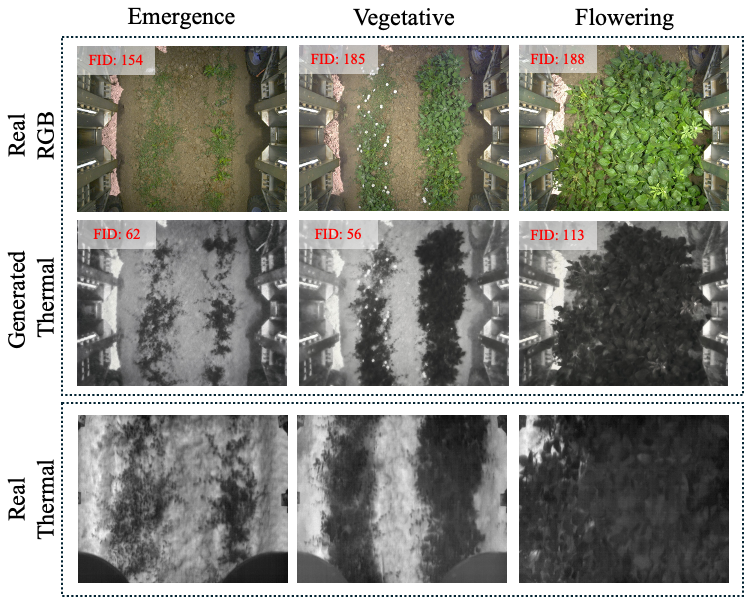}
    \caption{CycleGAN-turbo examples and calculated FID scores using the real thermal image as the target distribution.}
    \label{fig:fid_examples}
\end{figure}

\begin{figure}[h]
    \centering
    \includegraphics[width=0.8\linewidth]{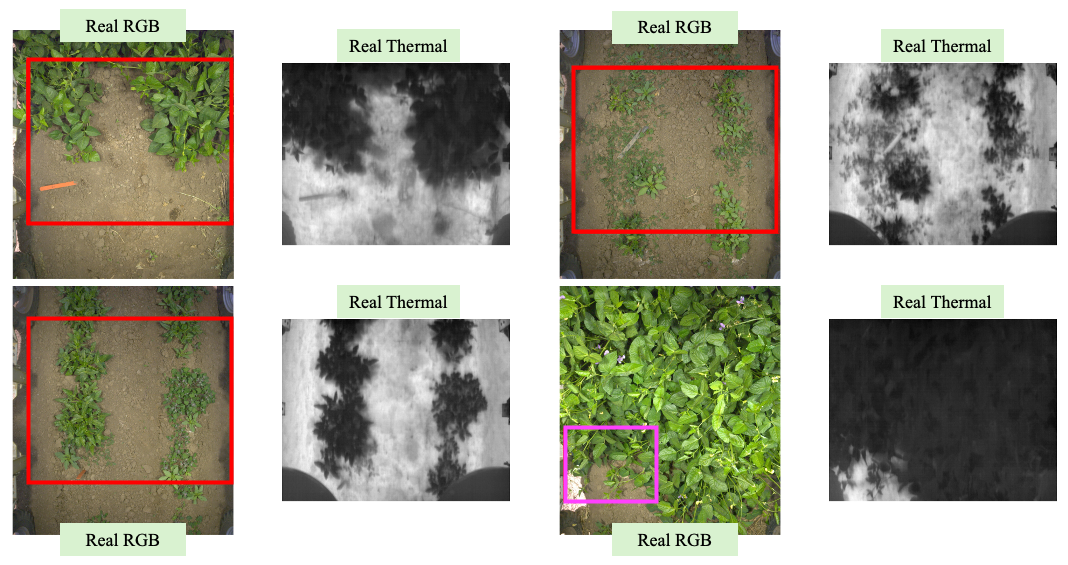}
    \caption{Template matching examples with red denoting successful matches and magenta denoting a failed match.}
    \label{fig:template_fails}
\end{figure}

By training solely on synthetic data, our best performing model is an FPN with an Efficientnet-b5 encoder, achieving a mean IoU of $0.565$ and mean Dice of $0.645$, as seen in Table \ref{tab:syn_results}. However, incorporating real images into the training set led to substantial performance gains. As shown in Figure~\ref{fig:syn_vs_real}, the mean IoU of a UNet++ model consistently improves with the addition of every five real images, reaching up to a 42\% relative increase compared to the synthetic-only baseline. We further evaluate per-class performance using the sample-based strategy and compare it to training on the full real dataset ($k=15$), as illustrated in Figure~\ref{fig:class_k}. Notably, the weed class (red) surpasses the full-data benchmark with just six to eight real images, while the plant class (blue) does so even earlier, with only two to four real images. When combining all synthetic and real data, we observed a maximum relative improvement of 22\% for weeds and 17\% for plants compared to the full real-data baseline. This improvement can be visually observed in Figure \ref{fig:sample_predictions}, where weed segmentation improves as $k$ number of images increases.

We also observed that incorporating just a small number of real samples, specifically five real images, can significantly boost model performance across architectures with varying scales and parameter counts. However, this improvement is not uniform: larger models with excessive parameter sizes may have diminished returns or even performance degradation. As shown in Figure~\ref{fig:model_k_5}, the SegFormer model with 86 million parameters experienced a decline in Dice score from 0.30 to 0.26 when real data was added. Similarly, the 86-million-parameter UNet++ model showed only a modest improvement, increasing from 0.10 to 0.18, suggesting that overparameterization may hinder effective learning when real data is limited.

\subsection{Cross-modal image generation and alignment}

We leveraged CycleGAN-turbo to synthesize thermal counterparts of our images, enabling accurate template matching against real thermal data. The Fréchet Inception Distance (FID) scores, shown in Figure~\ref{fig:fid_examples}, were computed using image subsets from each growth stage. While FID scores notably decrease during the emergence and vegetative stages, indicating improved realism in the translated thermal images, they rise again during the flowering and late-season stages. This suggests increasing translation difficulty as canopy complexity grows. Furthermore, we observed that template matching often fails when images are dominated by vegetation with few distinctive structural features, as illustrated in Figure~\ref{fig:template_fails}.

\section{Discussion and Limitations}

\subsection{Data scaling}

Training a segmentation model on over 1,000 synthetic images established a solid representational foundation of the domain. While, the synthetic only baseline already performs sufficiently well (see Figure \ref{fig:syn_vs_real} and \ref{fig:model_k_5}), we found that incorporating just a handful of real images, fewer than six, can yield noticeable gains in model performance. But, as we continuously incorporate labeled target data, the marginal influence of each additional real images diminishes, therefore proving that minimal manual labeling is required. These performance improvements are also partially influenced by the specific real images selected, which were curated to include a sufficient representation of both plant and weed classes. Additionally, the full set of real images spanned multiple phenological stages, suggesting that class diversity and temporal coverage both play roles in enhancing generalization.

Interestingly, models with a moderate number of trainable parameters consistently outperformed their larger counterparts. We hypothesize that in this relatively constrained binary segmentation task, lightweight architectures are less prone to overfitting on synthetic data. These smaller models may be better suited to integrating the small signal introduced by real data, perhaps due to stronger gradient responses or more stable optimization behavior. Over-parameterized models, on the other hand, may struggle to effectively utilize a sparse injection of real examples, especially when the domain shift remains significant.

Beyond this task, this domain adaptation strategy can definitely be applied to other applications and public datasets. One could pretrain on a similar dataset and then inject a handful of labeled images from the target domain to minimize the annotation effort while still reaching near-optimal performance. This can be further investigated to see how much more manual labels are needed for complex tasks. Or with a very large pretrained dataset (synthetic or not).

\subsection{Multi-modal alignment}

CycleGAN-Turbo enables RGB-to-thermal translation, facilitating cross-modal template matching for mask transfer. This approach is especially effective when plant features are clearly defined in both modalities. However, alignment performance degrades when thermal images are dominated by vegetation without distinctive structures, a common occurrence in dense canopies during mid-to-late season. This is expected, as thermal imagery suffers from lower spatial resolution and diminished texture detail compared to visible light. The absence of gradients and feature-rich regions in thermal views limits the reliability of pixel-wise correspondence. Nonetheless, the improved FID scores observed in early growth stages suggest that CycleGAN-Turbo effectively preserves modality-specific structure when alignment conditions are favorable.

This methodology could be expanded to other multi-modal imagery scenarios where the object of interest retains a recognizable structure, enabling robust template matching across modalities such as LiDAR or other multi-spectral data. In platforms equipped with multiple sensors, our alignment approach offers an alternative to purely timestamp-based synchronization, allowing computer vision algorithms originally developed for RGB inputs to be applied to new modalities. However, performance remains constrained when the scene lacks definitive visual features, for example, late-season canopies with heavy, uniform vegetation. So, future work should explore ways to consider correspondence under homogeneous conditions, perhaps by integrating temporal context, leveraging auxiliary sensor cues, or incorporating learned attention mechanisms to focus on even subtle structural cues.

\section*{Acknowledgments}

\subsection*{Author Contributions} 

\noindent Earl Ranario for method creation, evaluation and manuscript editing. \\
\noindent Brian N. Bailey for generating the synthetic images and manuscript editing. \\
\noindent Ismael Mayanja for proposing the research problem and manuscript editing. \\
\noindent Heesup Yun for contributing to alignment code and data collection. \\
\noindent J. Mason Earles for manuscript and methodology feedback.

\subsection*{Funding}

This work was supported by the Bill and Melinda Gates Foundation, Project ID: INV-002830, and USDA NIFA Hatch project 7003146. Under the grant conditions of the Foundation, a Creative Commons Attribution 4.0 Generic License has already been assigned to the Author Accepted Manuscript version that might arise from this submission.

\subsection*{Data Availability}
Synthetic and real datasets are available through AgML\footnote{https://github.com/Project-AgML/AgML} \cite{joshi_standardizing_2023}, a centralized framework for agricultural machine learning. AgML provides access to public agricultural datasets for common agricultural deep learning tasks, with standard benchmarks and pretrained models, as well the ability to generate synthetic data and annotations.

\printbibliography

\newpage

\section*{Supplementary Materials}
\renewcommand{\thefigure}{S\arabic{figure}}
\renewcommand{\thetable}{S\arabic{table}}
\setcounter{figure}{0}
\setcounter{table}{0}

\begin{figure}[h]
    \centering
    \includegraphics[width=1.0\linewidth]{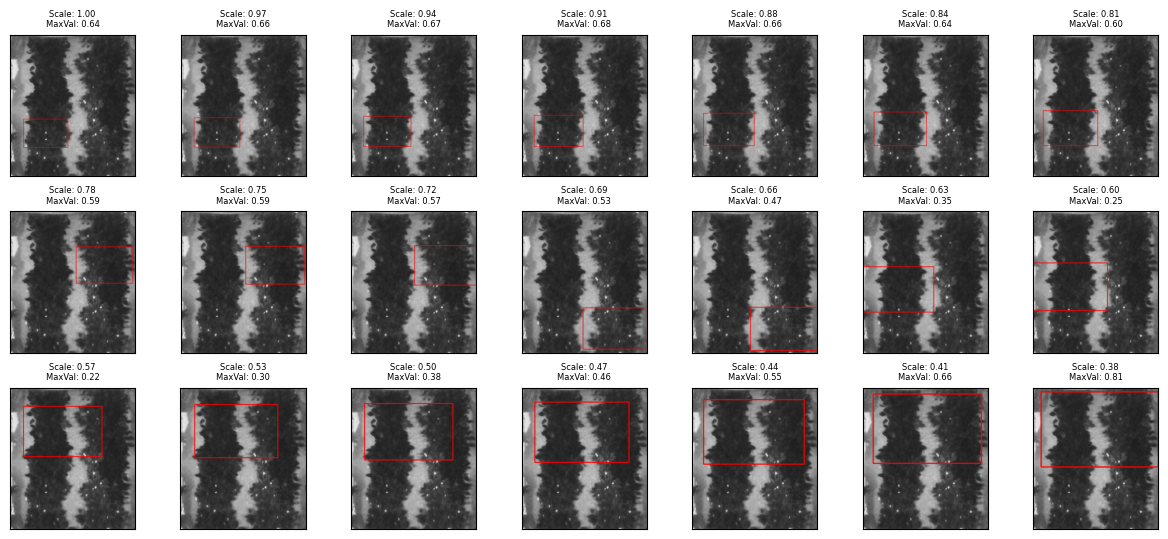}
    \caption{Template matching results across different scales.}
    \label{fig:template_matching}
\end{figure}

\begin{figure}[h]
    \centering
    \includegraphics[width=0.5\linewidth]{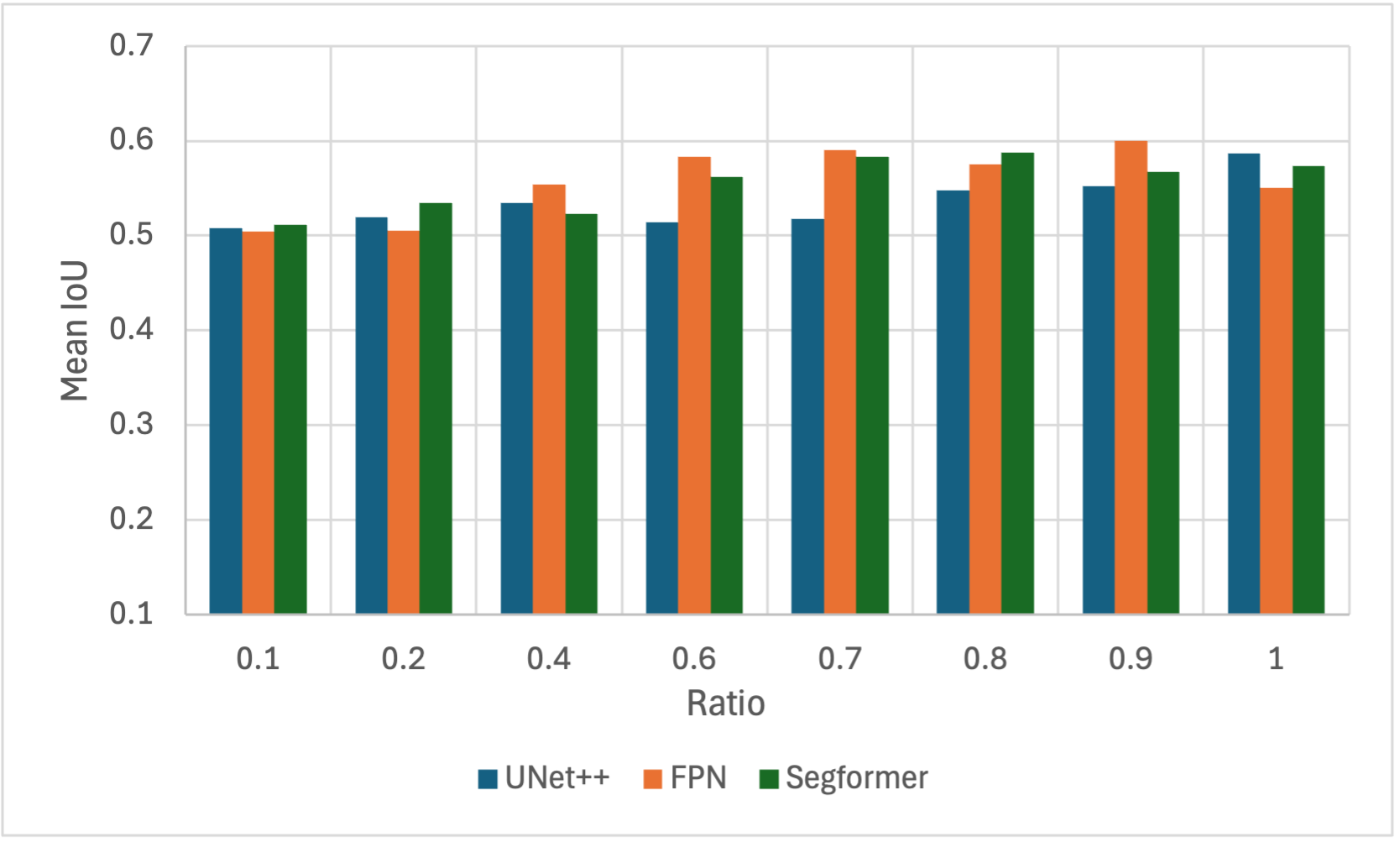}
    \caption{Test mean IoU scores for all model architectures (31M variants). Each score is evaluated at different ratios of the full synthetic dataset.}
    \label{fig:scaling_syn}
\end{figure}

\begin{figure}[h]
    \centering
    \includegraphics[width=0.5\linewidth]{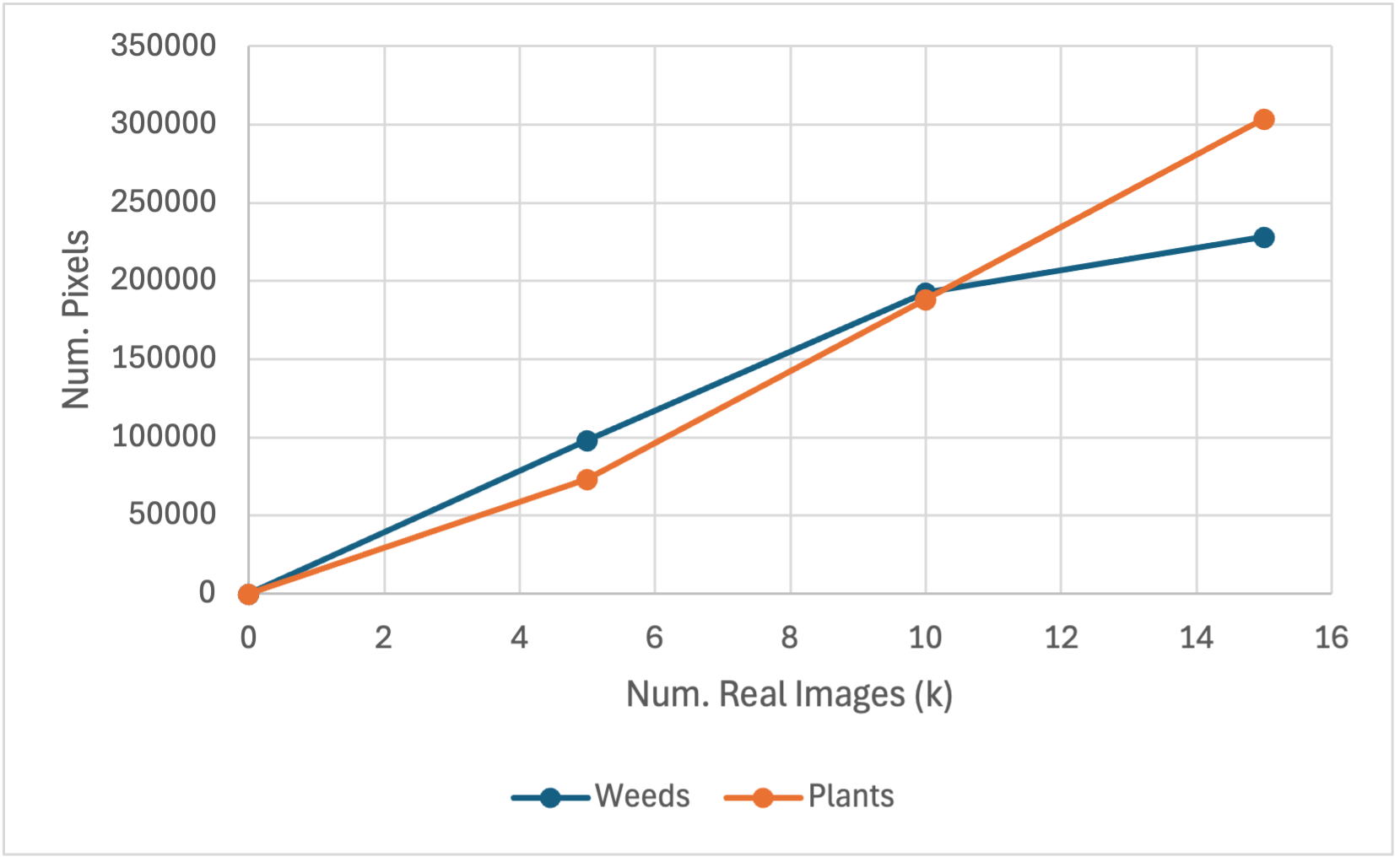}
    \caption{Number of pixels per class for each "k" selected images.}
    \label{fig:num_pixels_per_class}
\end{figure}

\begin{table}[h]
\centering
    \caption{Albumentations image augmentations used during training.}
    \label{tab:augmentations}
    \begin{tabular}{|p{4cm}|p{7cm}|c|}
    \hline
    \textbf{Augmentation} & \textbf{Function Call} & \textbf{Probability} \\ \hline
    Horizontal Flip & \texttt{A.HorizontalFlip(p=0.5)} & 0.5 \\ \hline
    ShiftScaleRotate & \texttt{A.ShiftScaleRotate(scale\_limit=0.5, rotate\_limit=0, shift\_limit=0.1, p=1, border\_mode=0)} & 1.0 \\ \hline
    PadIfNeeded & \texttt{A.PadIfNeeded(min\_height=imgsz, min\_width=imgsz, always\_apply=True)} & Always \\ \hline
    RandomCrop & \texttt{A.RandomCrop(height=imgsz, width=imgsz, always\_apply=True)} & Always \\ \hline
    GaussNoise & \texttt{A.GaussNoise(p=0.2)} & 0.2 \\ \hline
    Perspective & \texttt{A.Perspective(p=0.5)} & 0.5 \\ \hline
    \textbf{OneOf} & One of: CLAHE, RandomBrightnessContrast, or RandomGamma & 0.9 \\ \hline
    \quad CLAHE & \texttt{A.CLAHE(p=1)} & 1.0 \\ \hline
    \quad RandomBrightnessContrast & \texttt{A.RandomBrightnessContrast(p=1)} & 1.0 \\ \hline
    \quad RandomGamma & \texttt{A.RandomGamma(p=1)} & 1.0 \\ \hline
    \textbf{OneOf} & One of: Sharpen, Blur, or MotionBlur & 0.9 \\ \hline
    \quad Sharpen & \texttt{A.Sharpen(p=1)} & 1.0 \\ \hline
    \quad Blur & \texttt{A.Blur(blur\_limit=3, p=1)} & 1.0 \\ \hline
    \quad MotionBlur & \texttt{A.MotionBlur(blur\_limit=3, p=1)} & 1.0 \\ \hline
    \textbf{OneOf} & One of: RandomBrightnessContrast or HueSaturationValue & 0.9 \\ \hline
    \quad RandomBrightnessContrast & \texttt{A.RandomBrightnessContrast(p=1)} & 1.0 \\ \hline
    \quad HueSaturationValue & \texttt{A.HueSaturationValue(p=1)} & 1.0 \\ \hline
    \end{tabular}
\end{table}

\end{document}